\documentclass{egpubl}
\usepackage{VMV2025}
 
\WsPaper           %

\usepackage[T1]{fontenc}
\usepackage{dfadobe}  

\usepackage{cite}  %
\BibtexOrBiblatex
\electronicVersion
\PrintedOrElectronic
\ifpdf \usepackage[pdftex]{graphicx} \pdfcompresslevel=9
\else \usepackage[dvips]{graphicx} \fi

\usepackage{egweblnk} 

\usepackage{mathtools}
\usepackage{multirow}
\usepackage{amsmath}
\usepackage{microtype}
\usepackage{booktabs}
\usepackage[ruled,linesnumbered]{algorithm2e}
\usepackage[dvipsnames]{xcolor}
\usepackage[normalem]{ulem}

\definecolor{twred}{rgb}{0.85,0,0.2}
\definecolor{twgreen}{rgb}{0,0.65,0.2}

\def\clap#1{\hbox to 0pt{\hss #1\hss}}%
\def\initials#1{\protect\clap{\smash{\raisebox{1.4ex}{\tiny{\textsf{\textit{~~#1}}}}}}}%
\makeatletter
\newcommand{\NOTE}[3]{\protect\@ifundefined{hidecomments}{%
  \strut{\color{#2}{\hspace{0pt}\initials{#1}\protect{{\small$\lfloor$}#3{\small]}}}}%
  }{}}
\newcommand{\EDITbyauthor}[4][]{\protect\@ifundefined{hidecomments}{%
  \strut{\color{#3}{\hspace{0pt}\initials{#2}\protect\sout{#1}{#4}}}%
  }{}}
\newcommand{\EDITredandgreen}[4][]{\protect\@ifundefined{hidecomments}{%
  \strut{\color{twred}{\hspace{0pt}\protect\sout{#1}{\color{twgreen}{#4}}}}%
  }{}}
\newcommand{\EDITgreenonly}[4][]{\protect\@ifundefined{hidecomments}{%
  \strut{\color{twgreen}{#4}}%
  }{}}
\newcommand{\EDITfinal}[4][]{\protect%
  \strut{#4}%
  {}}

\newcommand{\EDIT}[4][]{\EDITfinal[#1]{#2}{#3}{#4}}

\newcommand{\NOTEboxed}[3]{\protect\@ifundefined{hidecomments}{%
  {\centering\fbox{\parbox{0.97\linewidth}{\protect\EDIT{#1}{#2}{#3}}}}%
  }{}}

\def\myhrulefill{\leavevmode\leaders\hrule height 2pt\hfill\kern\z@}

\makeatother

\newcommand{\TAedit}[2][]{\protect\EDIT[#1]{TA}{orange}{#2}}

\def\ignore#1{}

\usepackage{bm}

\usepackage{xspace}
\usepackage{pifont}%
\newcommand{\GG}[1]{}

\makeatletter
\DeclareRobustCommand\onedot{\futurelet\@let@token\@onedot}
\def\@onedot{\ifx\@let@token.\else.\null\fi\xspace}

\def\eg{\emph{e.g}\onedot} 
\def\ie{\emph{i.e}\onedot}

\def\etal{\emph{et al}\onedot}
\makeatother

\newcommand{\myparagraph}[2][\hspace{0.5em}]{\noindent\textbf{#2}#1}

\DeclareMathOperator{\argmin}{argmin}
\newcommand{\knn}{$k$-NN\xspace}

\usepackage[nolist]{acronym}
\begin{acronym}
    \acro{ad}[AD]{anomaly detection}
    \acro{ai}[AI]{artificial intelligence}
    \acro{al}[AL]{anomaly localization}
    \acro{aupro}[AUPRO]{area under the per-region overlap curve}
    \acro{auroc}[AUROC]{area under the receiver operating characteristic curve}
    \acro{cnn}[CNN]{convolutional neural network}
    \acrodefplural{cnn}[CNNs]{convolutional neural networks}
    \acro{emd}[EMD]{earth mover's distance}
    \acro{fca}[FCA]{feature correspondence analysis}
    \acro{fe}[$F$]{feature extractor}
    \acro{fpr}[FPR]{false positive rate}
    \acro{gan}[GAN]{generative adversarial network}
    \acrodefplural{gan}[GAN]{generative adversarial networks}
    \acro{hbos}[HBOS]{histogram-based outlier score}
    \acro{ms}{milliseconds}
    \acro{nn}[NN]{neural network}
    \acrodefplural{nn}[NN]{neural networks}
    \acro{pca}[PCA]{principal component analysis}
    \acro{pro}[PRO]{per-region overlap}
    \acro{qfca}[QFCA]{quantized feature correspondence analysis}
    \acro{roc}[ROC]{receiver operating characteristic}
    \acro{sc}{patch statistics comparison}
    \acro{svd}[SVD]{singular value decomposition}
    \acro{svg}[SVG]{scalable vector graphic}
    \acro{tpr}[TPR]{true positive rate}
    \acro{zsal}[ZSAL]{zero-shot anomaly localization}
    \acro{vlm}[VLM]{visual language model}
\end{acronym}

\title[Efficient Zero-Shot Texture Anomaly Detection]%
      {Quantized FCA: Efficient Zero-Shot Texture Anomaly Detection}

\author[A.-T. Ardelean \& P. Rückbeil \& T. Weyrich]
{\parbox{\textwidth}{\centering Andrei-Timotei Ardelean\orcid{0000-0001-9317-5149}
        and Patrick Rückbeil\orcid{0009-0002-0018-0673}
        and Tim Weyrich\orcid{0000-0002-4322-8844} 
        }
        \\
        \parbox{\textwidth}{\centering Friedrich-Alexander-Universität Erlangen-Nürnberg, Germany
        }
}

\begin{document}

\maketitle
\begin{abstract}
    Zero-shot anomaly localization is a rising field in computer vision research, with important progress in recent years. This work focuses on the problem of detecting and localizing anomalies in textures, where anomalies can be defined as the regions that deviate from the overall statistics, violating the stationarity assumption. The main limitation of existing methods is their high running time, making them impractical for deployment in real-world scenarios, such as assembly line monitoring. We propose a real-time method, named QFCA, which implements a quantized version of the feature correspondence analysis (FCA) algorithm. By carefully adapting the patch statistics comparison to work on histograms of quantized values, we obtain a 10$\times$ speedup with little to no loss in accuracy. Moreover, we introduce a feature preprocessing step based on principal component analysis, which enhances the contrast between normal and anomalous features, improving the detection precision on complex textures. Our method is thoroughly evaluated against prior art, comparing favorably with existing methods. \\
    Project page: {\footnotesize\texttt{\href{https://reality.tf.fau.de/pub/ardelean2025quantized.html}{reality.tf.fau.de/pub/ardelean2025quantized.html}}}.


\end{abstract}

\renewcommand{\pageref}[1]{%
  \ifstrequal{#1}{ThisPartLastPage}{10}{\ref{#1}}%
}

\section{Introduction}
\label{chap:intro}
The task of finding outliers in a group of otherwise similar elements is called \ac{ad}. These outliers are also referred to as anomalies, discordant observations, exceptions, aberrations, surprises, peculiarities, or contaminants depending on the particular context \cite{anomalyDetection_survey}. 
Some notable fields of application are manufacturing inspection \cite{Liu2024, Hsieh2019js}, medical imaging \cite{fernando2021deeplearningmedicalanomaly}, as a preprocessing step in data analysis \cite{dataaugexperiment, Smith2011-jt}, in machine vision \cite{Aggarwal2016-oi} and neuroscience \cite{Bibi2021-ku}, in weather records \cite{Wibisono2021-ie}, for fraud detection in the financial industry \cite{Ahmed2016-tp}, as well as monitoring acoustic \cite{useAcoustic} or video \cite{Kaur2018-tt} signal. 
In most use-cases of image AD, it is crucial to not only detect if anomalies are present, but also segment the offending regions.
The task of \ac{zsal}, targeted by this work, refers to locating such outliers without prior training or an indication as to what outliers or even non-outlier regions might look like.

\begin{figure}[!t]
	\centering
	\includegraphics[width=0.99\columnwidth]{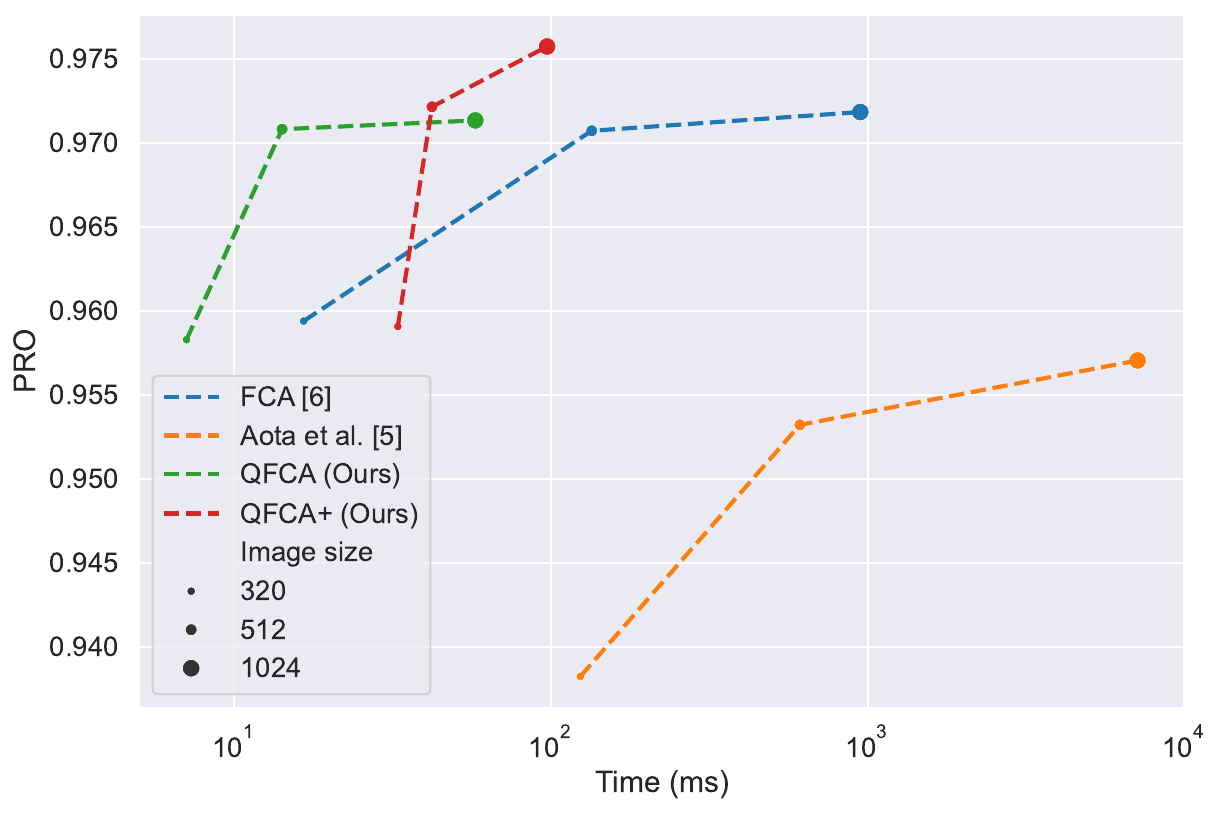}
	\caption{\label{fig:teaser}Anomaly localization fidelity (PRO) vs Time, obtained using different algorithms at various resolutions. Our approach enables the best tradeoff between accuracy and runtime. 
}
\end{figure}

Humans are generally capable of making the distinction between normal elements and irregularities, when given a set of elements without labeled outliers \cite{Tao_2022}. 
Given some form of homogeneous data, like textures, anomalies would arise in the form of disturbances of the homogeneity. 
This assumption of homogeneity, or stationarity, supports the idea of automatically locating anomalies in a single image, based on the internal statistics.

In this paper, we approach the problem of zero-shot anomaly localization with a specific focus on textures.
The method builds upon \ac{fca}~\cite{ardelean2023highfidelityArxiv}, the current state-of-the-art algorithm for this task, lifting its crucial limitation: running time.
We identify the most time-consuming operations in FCA and replace them with a quantized version, obtaining a significant efficiency boost (one order of magnitude) with a limited decrease in accuracy.
To further minimize the running time, we employ an efficient algorithm for local average pooling that runs in constant time with respect to kernel size.
Moreover, we found the \knn version of \ac{fca} (designed to improve anomaly localization on more complex textures) prohibitively slow for images larger than 320 \texttimes{} 320. To mitigate this issue, we advance a feature preprocessing technique that yields a similar improvement in accuracy with a much lower running time.

In essence, our contributions are as follows:
\begin{itemize}
    \item We introduce a real-time algorithm for zero-shot anomaly localization with unprecedented detection accuracy.
    \item We propose a feature preprocessing technique that improves detection on complex textures.
    \item We identify and resolve a computational bottleneck caused by an inefficient implementation of local average pooling in modern ML libraries, including PyTorch, TensorFlow, and JAX.
\end{itemize}
    
\section{Related Work}

The field of \ac{ad} has a rich research history and a wide range of applications, with various methods being thereby specialized for different contexts and conditions.
We refer the reader to the survey of Hodge and Austin \cite{Hodge2004-ba} for a wide-scoped exposition about different approaches to outlier detection and to Liu \etal~\cite{liu2024deep} for a modern outlook on Visual Anomaly Detection.

\myparagraph{Unsupervised anomaly detection.}
The most common setting for \ac{ad} in images is unsupervised anomaly detection. More correctly referred to as one-class classification or normality-supervised~\cite{ardelean2024blind}, the methods in this category use a curated set of \emph{normal} images, guaranteed not to contain anomalies.
An early popular approach for \acl{ad} using \acp{nn} includes reconstruction-based methods~\cite{Tao_2022}, which try to recreate the image content from limited information or a bottleneck. This can be achieved using (variational) autoencoders \cite{variationAutoencoder-jinwon, conf/visapp/BergmannLFSS19, defectSegmentationSteelStripSurface, baur2019deep} or \acp{gan} based approaches such as Skip-GANomaly \cite{Akcay_2019} and SAC \cite{yang2022transformer}.
During inference, these methods localize anomalies using the difference between the original and the reconstruction. The underlying intuition is that the model fails to reconstruct anomalous regions since it was trained only on \emph{normal} images.

To achieve high-fidelity detections, most recent approaches leverage neural features extracted with a CNN pretrained on ImageNet~\cite{iamgenet}. Various methods make use of these features in different ways, such as computing the $k$-nearest neighbors (\knn)~\cite{Bergman2020-journals/corr/abs-2002-10445, cohen2021subimageanomalydetectiondeep}, modeling multivariate Gaussians at the pixel level~\cite{defard2021padim}, creating a memory bank~\cite{patchcore}, and contrastive supervision~\cite{lee2022cfa}.
A way to speed up feature extraction is to train a smaller network to predict the teacher's features~\cite{batzner2024efficientadaccuratevisualanomaly}.

A significant category of unsupervised anomaly detection is represented by the \textbf{few-shot} setting. A growing number of works tackle the problem of anomaly detection with small amounts of \emph{normal} data~\cite{huang2022registration, lee2023few, li2024promptad, huang2024few}. These methods try to maximize the information extracted from a handful of images and incorporate cross-category knowledge from other image classes. 
Although the few-shot setting is significantly more challenging compared to the general case, there is still a large difference to the zero-shot setting addressed in our work, where no reference images are provided.

\myparagraph{Zero-shot anomaly detection.} 
Localizing anomalies from a single image of a previously
unseen image class is the most challenging setting in AD, as it requires detecting anomalies using a small amount of contaminated data. 
In this sense, it is similar to blind anomaly detection~\cite{zhang2024s, ardelean2024blind} at the level of image patches, albeit with a smaller sample size.
The task was formalized in MAEDAY \cite{schwartz2024maeday}, where a zero-shot solution was proposed, based on a masked autoencoder (MAE).
The works that followed can be generally grouped into two categories: based on a \ac{vlm} or based on internal statistics.

To contend with the lack of supervisory signal in the zero-shot setting, \ac{vlm}-based methods seek to leverage the massive pretraining of these models to distinguish between normal and anomalous appearance.
WinCLIP~\cite{jeong2023winclip} is the pioneer of this approach to \ac{zsal}; it identifies anomalies through carefully designed prompts, running the CLIP~\cite{radford2021learning} model on several scales to obtain localized patch-prompt matching scores.
Several methods have since been proposed that improve upon the idea of WinCLIP by bridging the domain gap (April-GAN)~\cite{chen2023zero}, using object-agnostic prompts (AnomalyCLIP)~\cite{zhou2023anomalyclip}, leveraging SAM for mask proposals (SAA)~\cite{cao2023segment}, improving the computation of representative text features (SDP)~\cite{chen2024clip}, and combining CLIP and SAM (ClipSAM)~\cite{li2025clipsam}.

The second category is suitable for textures and relies on the information within the image to model the distribution of \emph{normal} features. The anomalies are therefore the regions that are outliers with respect to the overall content of the texture.
The method developed by Aota \etal \cite{Aota_2023_WACV} (denoted ZvM) builds on the idea that an anomalous feature patch has fewer similar patches across the image compared to a normal one. As such, the mean distance to the k-nearest neighbors for a suitable $k$ will be significantly higher for anomalous patches, enabling their localization.

Ardelean and Weyrich~\cite{ardelean2023highfidelityArxiv} propose a framework for \ac{zsal} consisting of three components: feature extraction, reference selection, and patch statistics comparison. FCA itself is an implementation of this abstraction that provides high-fidelity detections by using a fine-grained error-to-pixel association.
The algorithm creates patches for each spatial location and compares them with a global representation of the texture. The comparison computes a patchwise error map using a bijective mapping derived from the 1-dimensional Wasserstein Distance \cite{elnekave2022generating}.
Lastly, a score is computed using the aggregation of the pixel-wise contribution to the error maps of all nearby patches, which are finally averaged across the channel dimension (more details in Section \ref{subsec:fca_sc}).

\section{Algorithm Design}
\label{chap:algorithm}
Our method, named \ac{qfca}, follows the framework of FCA for zero-shot localization. That is, the anomaly maps are computed according to the following formula:
\begin{align}
    A(x, y; F,S,R) := \sum_{\mathclap{F_r \in R(F(I))}}{S(x, y, F(I), F_r)}\;,
\end{align}
where $F$ denotes feature extraction, $S$ compares the patch statistics, and $R$ computes a set of global references that characterize the texture. 
Informally, the equation defines the anomaly score at a pixel location $x, y$ as the sum of errors associated with that pixel when comparing the features of the surrounding patch with the set of global references.
Our method differs from FCA in the implementation of these three components, which will be described in the following subsections.

\subsection{Patch Statistics Comparison}
\label{subsec:fca_sc}

The patch statistics comparison function $S(x,y,F(I),F_r)$ evaluates the degree of anomaly for each position $(x,y)$ within its local context in the feature map \TAedit{$F(I) \in \mathcal{R}^{C\times H\times W}$}. To do so, the function compares the local statistics around $(x,y)$ and the statistics of a global reference $F_r$.

\begin{figure}
	\centering
	\includegraphics[width=0.99\columnwidth]{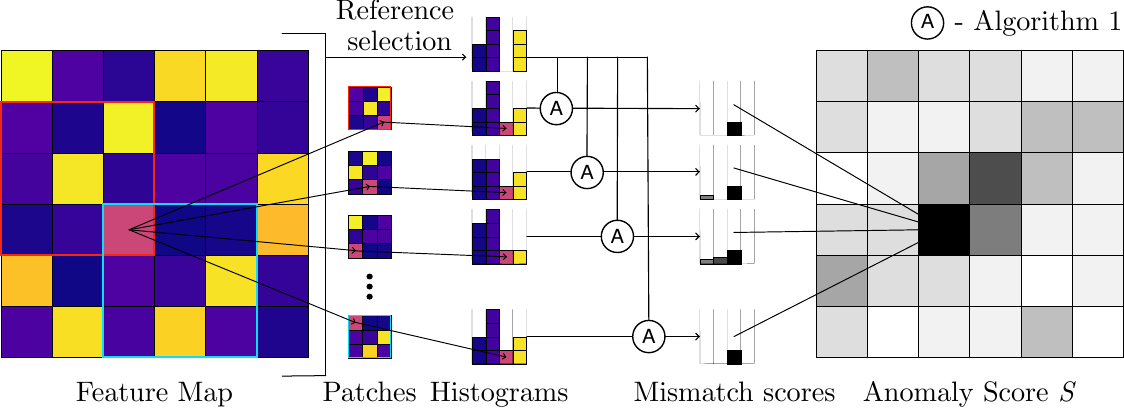}
	\caption{Our QFCA pipeline: each patch is represented as a histogram and compared to the reference using our Algorithm \ref{algorithm}. The final score for each pixel is given by averaging the mismatch scores from the histograms it is part of.}
	\label{fig:qfca_pipeline}
\end{figure}

To keep the description self-contained, we begin by detailing the patch statistics comparison of FCA, and then present our approach:
FCA first separates the image into multiple overlapping patches by taking each pixel and its surroundings. 
For each patch $P$, a bidirectional mapping is formed between the pixels and the reference by sorting the feature values and mapping elements of the same rank.
The error associated with each pixel $(x,y)$ is the absolute difference between the feature value and the corresponding element in the reference.
Formally, this error represents the contribution of the pixel to the optimal transport between the empirical distribution of the patch and the reference.
This is repeated for each feature channel independently and aggregated into a summative mismatch score $M(x,y; P)$ for each pixel of that patch $P$.

The most important design decision of the algorithm is that the anomaly score for a specific location is not computed based on the mismatch scores of nearby pixels, as in ZvM~\cite{Aota_2023_WACV}; instead, it is computed by aggregating the mismatch scores of that pixel in the context of the nearby patches (of which it is part).
Formally, FCA defines it as

\begin{align}
\label{eq:fca_eq}
    S(x,y) = \smashoperator{\sum_{(x',y')\in P_{xy}}} \mathcal{G}_{\sigma_s}(M(\cdot; P_{x'y'}))(x,y) G_{\sigma_p}(x'-x,y'-y) ,
\end{align}

where $\mathcal{G}_{\sigma_s}$ denotes Gaussian blurring with standard deviation $\sigma_s$ and $G_{\sigma_p}(x, y)$ directly retrieves the PDF of the bivariate normal distribution.

The most time-consuming operation of FCA is represented by the sorting operation that creates the bidirectional mapping between elements of the same rank. This is especially costly, as it must be performed for each patch for each channel.
The main quality of our QFCA (Fig.~\ref{fig:qfca_pipeline}) is the quantized representation of patches as histograms.
As shown in our experiments (Sec.~\ref{subsec:bin_nr}), by carefully adapting the error computation and association, we obtain a localization performance similar to that of the non-quantized version with as little as 16 bins.

In the following, we describe our efficient algorithm for comparing a patch to a global reference and computing the error contribution analogous to the operation in FCA. In the supplementary, we provide a proof for the correctness of the algorithm and a mathematical justification of its efficacy based on its relation to the gradient of the 2-Wasserstein distance (\ref{ssec:correct}).

Firstly, we quantize the feature values into $N$ equally-spaced values $\{Q_i\}_{i=1}^N$. Each patch is then represented using histograms based on the quantized feature values; we denote with $\{P_i\}_{i=1}^N$ and $\{R_i\}_{i=1}^N$ the histogram weights of the patch and reference vectors, respectively. 
Since we deal with empirical distributions, the histogram weights are integers; however, the following algorithm can be applied unchanged to arbitrary weights.

\noindent To find the optimal transport, we use a two-pointer ($i, j$) algorithm that iterates the matching quantiles in $P$ and $R$.
At each step, the error attributed to the current bin in $P$ is given by $\min(P_i, R_j) \cdot |Q_i - Q_j|$, indicating a transport between bin $i$ and bin $j$. Then, the index pointing to the smaller weight is moved forward, while the amount of mass that was transported, \ie $\min(P_i, R_j)$, is removed from the larger bin. See Algorithm \ref{algorithm} for a formal description.
\SetKwFor{RepTimes}{repeat}{times}{end}
\begin{algorithm}[!ht]
    \caption{\label{algorithm}Quantized patch mismatch score}
    \SetAlgoLined
    \KwIn{$\{P_i\}_{i=1}^N$, $\{R_i\}_{i=1}^N$, $\{Q_i\}_{i=1}^N$}
    Initialize mismatch scores $E_i \gets 0,\; i=1..N$ \\
    Make a copy of $\{P_i\}_{i=1}^N$ in $\{\hat{P}_i\}_{i=1}^N$ \\
    $i \gets 1$, $j \gets 1$\;
    \While{$i \le N$ and $j \le N$}{
        \uIf{$P_i < R_j$}{ 
            $E_i \gets E_i + P_i \cdot |Q_i - Q_j|$ \;
            $R_j \gets R_j - P_i$ \;
            $i \gets i+1$ \;
        }
        \Else {
            $E_i \gets E_i + R_j \cdot |Q_i - Q_j|$ \;
            $P_i \gets P_i - R_j$ \;
            $j \gets j+1$ \;
        }
    }
    $E_i \gets \frac{E_i}{\hat{P_i}}$ \textbf{for} $i = 1..N$\;
    \Return $\{E_i\}_{i=1}^N$
    \label{alg:sampling}
\end{algorithm}

\TAedit{After completion of the algorithm, $E_i$ stores the contribution of the bin $i$ to the Wasserstein distance between the patch $P$ and the reference, which is the histogram equivalent of the FCA mismatch score $M(x, y; P)$ for a certain channel.}
The division on line 15 ensures that the mismatch score for each bin is weighted by the number of elements that contributed to the error.
It can be easily seen that the algorithm finishes after $2N-1$ steps and therefore has a linear complexity in the number of bins compared to the $O(T^2\log(T))$ of FCA (where T is the patch size). Moreover, we compile the code to a CUDA kernel and benefit from GPU acceleration, as the computation can be performed in parallel for all patches and all channels.

The error-to-pixel association \TAedit{(Eq. (\ref{eq:fca_eq}))} is a crucial component of the algorithm, which differentiates it from a simple Wasserstein distance between the patches and the reference. 
\TAedit{In order to recover the same formulation as FCA, }after computing the bin-level error for each patch, we distribute the error contribution to all pixels that formed each histogram bin.
Similarly to the Gaussian smoothing from Eq. (\ref{eq:fca_eq}), we aggregate the errors of all histogram bins of which a certain pixel is part. We simply implement this operation as a Gaussian blurring over the bin-level errors with a standard deviation of $\sigma_p$ and a kernel size equal to the patch size.
Finally, the anomaly score of a pixel is given by its corresponding bin after smoothing.

\begin{figure}
	\centering
	\includegraphics[width=0.99\columnwidth]{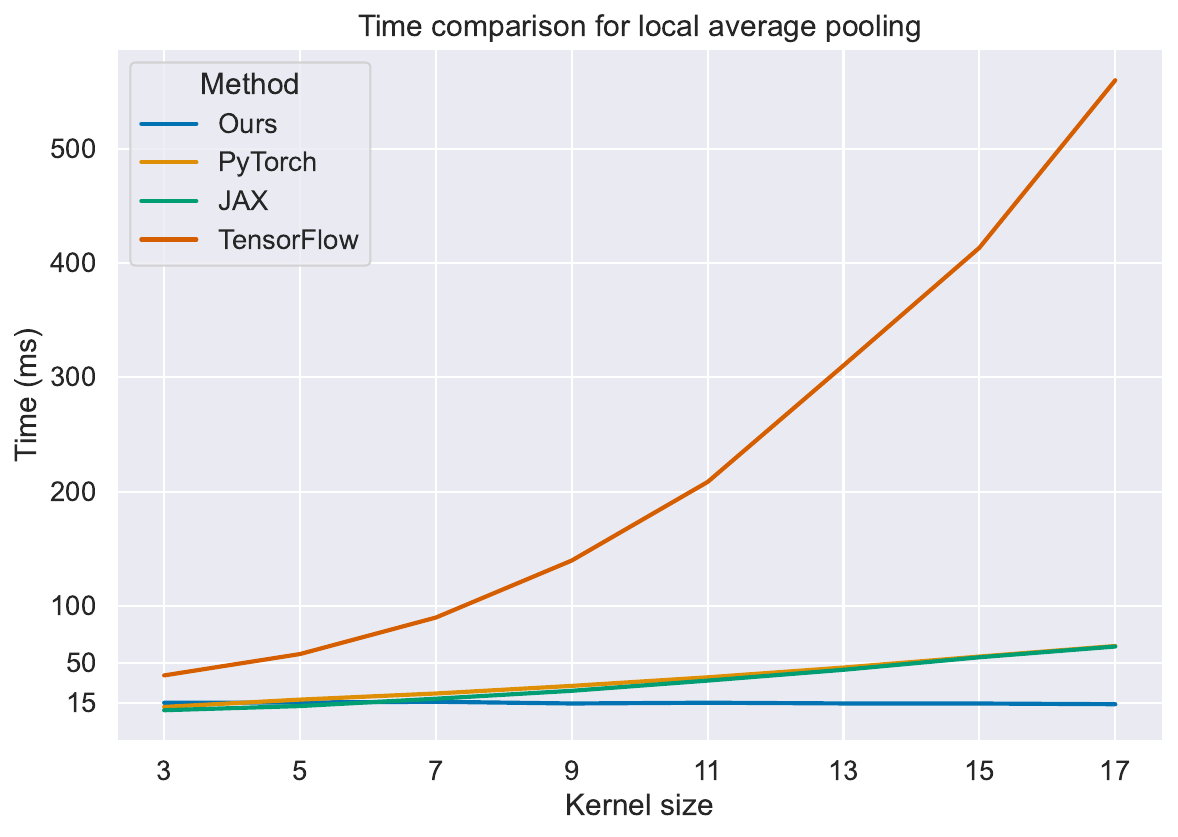}
	\caption{Runtime for average pooling with different kernel sizes, operating on a 128\texttimes{}128 tensor with 8192 channels (mirroring our 16 bins per feature setting). All methods use GPU acceleration.}
	\label{fig:avg_pooling}
\end{figure}

As highlighted in the description of our algorithm, its time complexity does not depend on the patch size $T$; however, our initial experiments still revealed a linear dependence with $T$ of the overall running time. More precisely, the patch-size dependence occurred in two places: the formation of histograms and the error-to-pixel association, even as these operations were implemented as local average pooling.
To our surprise, we found that the running time of the PyTorch implementation of average pooling increases with the kernel size; upon inspection, the same observation was made for other major ML frameworks, \ie, TensorFlow and JAX.
To address this, we implement the 2D average pooling using summed-area tables~\cite{crow1984summed} (also known as integral image~\cite{viola2001rapid}), having an $O(H\times W)$ complexity per channel.
The algorithm readily benefits from parallelization, resulting in a fast runtime, as shown in Figure~\ref{fig:avg_pooling}.
Using our fast average pooling implementation, we reduce the overall running time of our method by 30\% in the usual case (patch size of 9) and make our algorithm's complexity completely independent of the patch size.

\subsection{Reference Selection}
\label{chap:fca_ref}

Ardelean and Weyrich~\cite{ardelean2023highfidelityArxiv} explored various options for computing the set of references $R(F(I))$, such as: all patches, random selection, patch average, and median.
The latter proved to be the best option in terms of detection fidelity and running time,
\TAedit{which is because it uses a single reference to capture the entire global statistics $R(F(I)) = \{F_r\}$.}
\ac{fca} defines the feature reference $F_r$ as:
$$\underset{F_r}{\argmin} \smashoperator{\sum_{(x,y)}} A(x, y; \cdot, \cdot, R={F_r})\;, $$
which is the distribution that minimizes the Wasserstein distance across all patches. \TAedit{This minimization problem has a closed-form solution}, obtained by computing the median over the patches for each rank in sorted order. A reference vector is computed for each feature channel separately.

We similarly use median-based reference selection by adapting it to our quantized representation of features. We find that the best results are obtained when using the same method as \ac{fca} to compute the median with full precision and only quantize the reference afterward. Section \ref{chap:reference} compares different ways to compute the reference set in terms of accuracy.

Although the median reference computation works well for homogeneous images, it struggles with complex textures \eg, where the texture period is larger than the patch size. An idea that was investigated in \ac{fca} is a pairwise comparison with the $k$-nearest neighbors of all patches. This showed a clear improvement in accuracy for elaborate textures, albeit with a severely slower runtime.
Our paper introduces an alternative approach that increases the accuracy for more complex images, without making the method prohibitively slow. Our approach enhances the features by reducing the variance of normal, frequently-occurring features, as described in the following section.

\subsection{Feature Extractor}
\label{chap:fca_fe}

The task of the \ac{fe} is to create spatial features based on the input image. Virtually all zero-shot anomaly detection methods based on statics rely on medium- and high-level descriptors, which are more suitable than raw pixels to differentiate between normal and anomalous appearance.
Pretrained CNNs have been shown to provide features that are better than hand-annotated features \cite{deepfeature}.
Specifically, Roth \etal~\cite{patchcore} evaluated different neural networks and found Wide ResNet-50~\cite{zagoruyko2017wideresidualnetworks} to perform especially well on the task of anomaly detection.
Considering the good results and acceptable running times of this CNN, we also use it for feature extraction in our pipeline.

Inspired by the use of VAE residuals in BlindLCA~\cite{ardelean2024blind}, we propose a single-image feature preprocessing technique with the aim of reducing the variance of normal features in the case of intricate images.
For example, in bimodal textures, where the period of the texture is larger than the patch size, a single global reference cannot capture the entire complexity.
\Ac{fca} + \knn aims to fix this by computing all pairwise distances between patches and then discarding outliers by considering only the closest $k$ distances. Nonetheless, this method has a high running time, as it scales quadratically with the number of patches \cite{ardelean2023highfidelityArxiv}.
Conversely, we build on the idea behind reconstruction-based methods, and use the residuals between the original image and the reconstruction as the new features on which we apply QFCA.
Differently from BlindLCA, we use a single image and employ principal components analysis (PCA) for reconstruction in order to preserve a fast running time.
Intuitively, our feature preprocessing follows the idea that anomalies will not be well represented by the principal feature components of the input features. Therefore, subtracting the PCA reconstruction from the input reduces the variance of normal features, while preserving that of anomalies. This technique helps with elaborate textures, for which the reference struggles to capture the global statistics (Sec.~\ref{chap:fca_ref}).

Formally, our improved feature extractor is defined as:
\[ F(X) = W(X) - \text{PCA}^{-1}(\text{PCA}(W(X))) \;,\]
where $W$ represents the Wide Resnet-50 network; \ie, the feature extracted after the second convolutional block. We denote our full method, which uses this preprocessing, as QFCA+.
The number of principal components used in dimensionality reduction plays an important role, which we analyze in Section~\ref{chap:feature_enhancer}.

\section{Experiments}
\label{chap:experiments}

We evaluate our QFCA algorithm and the feature preprocessing step on three different datasets using the established metrics. 
The experiments show that our approach localizes anomalies faster and more accurately compared to existing zero-shot methods.

\subsection{Datasets}
\label{chap:datasets}

\noindent\textbf{MVTec AD}~\cite{mvtec2019, Bergmann2021-xo} is the current industry standard dataset for anomaly localization.
As our method is designed to work on textures, we use the five texture classes: carpet, grid, leather, tile, and wood; together they accumulate over 500 images and their respective segmentation masks.

\noindent\textbf{DTD-Synthetic} is a synthetic dataset by Aota \etal~\cite{Aota_2023_WACV}, created for evaluating anomaly detection on more diverse and complex textures. It is based on DTD (Describable Texture Dataset) \cite{Cimpoi2014textures}, and consists of 12 textures with 100 training images and over 100 test images each. 
The images are cropped out of the original DTD image with a random orientation and position, resulting in image sizes between 180\texttimes{}180 and 384\texttimes{}384 pixels.

\noindent\textbf{Woven Fabric Textures} (WFT) introduced by Bergmann \etal \cite{conf/visapp/BergmannLFSS19} is a dataset with 2 textures with 50 images each.
The dataset includes ground-truth segmentation masks at a resolution of 512\texttimes{}512.

\subsection{Metrics}
\label{chap:metrics}

For the evaluation of the results, we use the common metrics and the same post-processing steps as in \ac{fca}. That is, the borders are discarded and the image-level anomaly score is computed as the maximum of the pixel-level scores.

Anomaly localization is customarily evaluated using the AUROC$_s$ \cite{ROC-fogarty} and the \ac{aupro}~\cite{Bergmann2021-xo} metrics. Both metrics are threshold independent and operate at the pixel level. 
The PRO score was introduced to counteract the bias toward large anomalies, which is done by weighing the anomaly predictions by the area of the anomalous region in the ground truth. 
As recommended by Bergmann \etal \cite{mvtec2019}, we compute the area under the PRO curve up to a false positive rate of 30\%; for brevity, we simply refer to this metric as PRO in the rest of this paper.
Additionally, we quantify the anomaly localization fidelity at a fixed threshold using the $F_1$ metric~\cite{chinchor1992muc}, \ie, the harmonic mean between precision and recall calculated at the optimal threshold.
As the purpose of this work is to optimize the running time and enable real-time usability, we also report the execution time for one image (s / img). Note that we measure latency, rather than throughput, keeping in mind a live-feed use-case.

As acknowledged by the community~\cite{zhang2024learning}, image-level AUROC$_c$ is saturated on datasets such as MVTec (scores above $99.5\%$). Therefore, we use the more representative pixel-level metrics in the main paper and refer the reader to the supplementary (\ref{asec:detailed_metrics}) for extended numerical results.

\subsection{Comparison with prior art}

\begin{figure*}[ht]
\setlength{\tabcolsep}{1pt}
\renewcommand{\arraystretch}{0.7}
\newlength{\imw}
\setlength{\imw}{0.13\textwidth}
\centering
\begin{tabular}{@{}c@{\hspace{2mm}}c@{\hspace{1mm}}c@{\hspace{1mm}}c@{\hspace{1mm}}c@{\hspace{1mm}}c@{\hspace{1mm}}c@{\hspace{1mm}}c@{}}

\multirow{2}{*}[2em]{\rotatebox[origin=c]{90}{ MVTec AD}} %
 & \includegraphics[width=\imw]{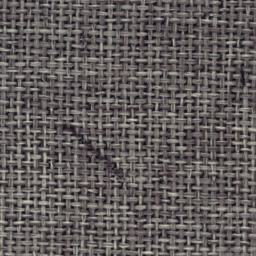} &
\includegraphics[width=\imw]{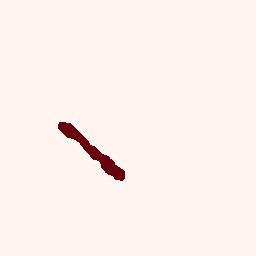} &
\includegraphics[width=\imw]{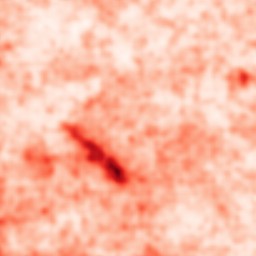} &
\includegraphics[width=\imw]{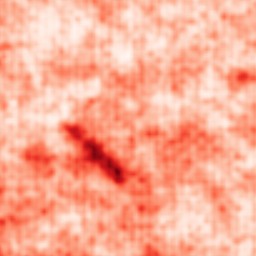} &
\includegraphics[width=\imw]{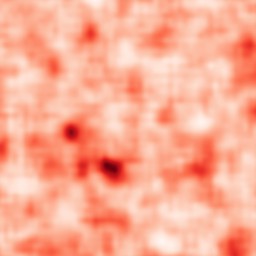} &
\includegraphics[width=\imw]{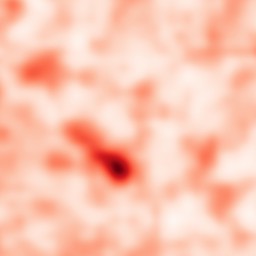} &
\includegraphics[width=\imw]{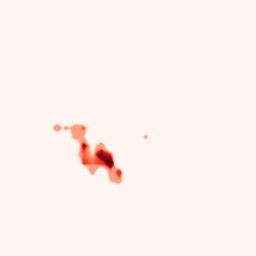} \\

& \includegraphics[width=\imw]{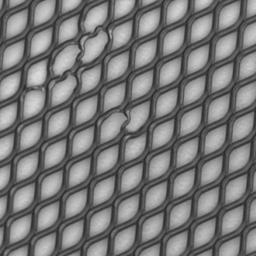} & %
\includegraphics[width=\imw]{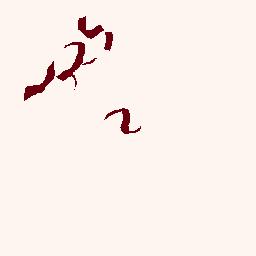} &
\includegraphics[width=\imw]{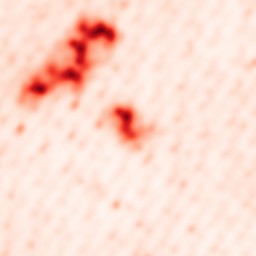} &
\includegraphics[width=\imw]{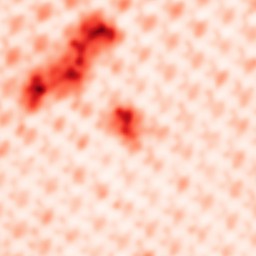} &
\includegraphics[width=\imw]{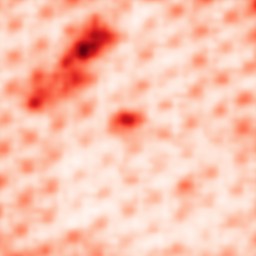} &
\includegraphics[width=\imw]{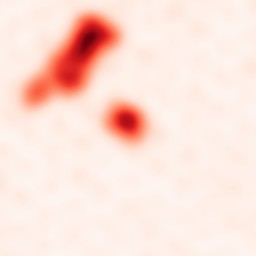} &
\includegraphics[width=\imw]{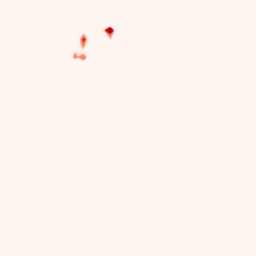} \\

\multirow{2}{*}[2em]{\rotatebox[origin=c]{90}{DTD-synthetic}} %
& \includegraphics[width=\imw]{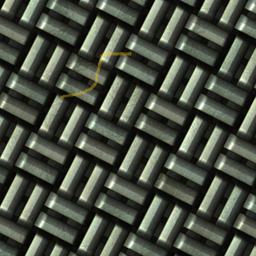} &
\includegraphics[width=\imw]{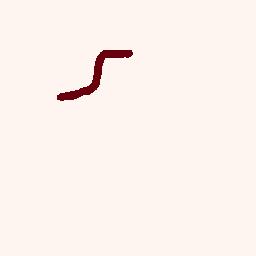} &
\includegraphics[width=\imw]{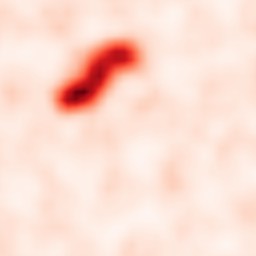} &
\includegraphics[width=\imw]{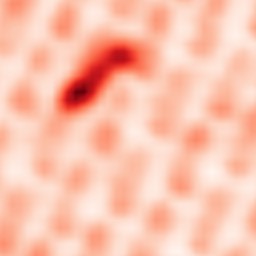} &
\includegraphics[width=\imw]{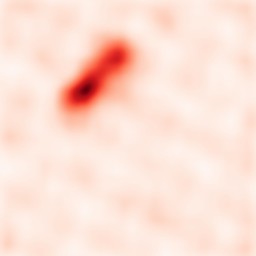} &
\includegraphics[width=\imw]{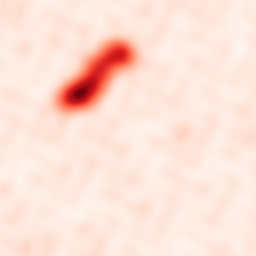} &
\includegraphics[width=\imw]{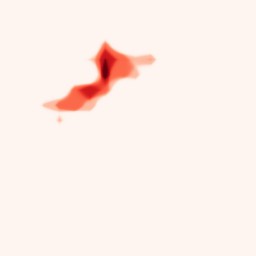} \\

& \includegraphics[width=\imw]{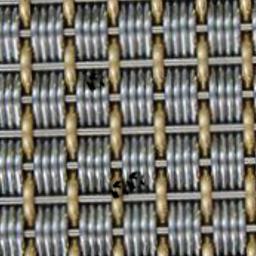} & %
\includegraphics[width=\imw]{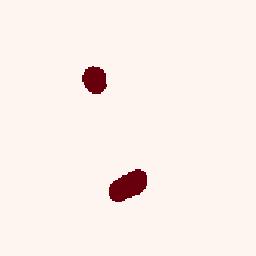} &
\includegraphics[width=\imw]{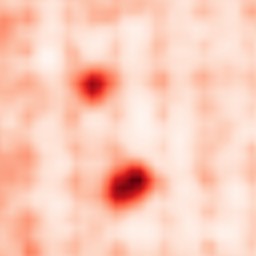} &
\includegraphics[width=\imw]{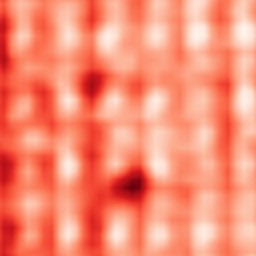} &
\includegraphics[width=\imw]{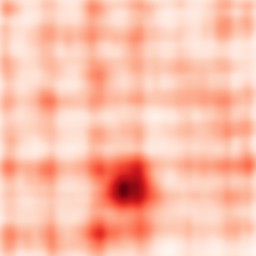} &
\includegraphics[width=\imw]{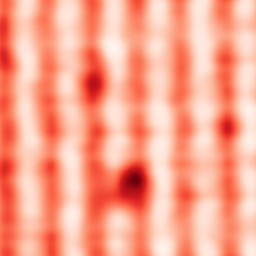} &
\includegraphics[width=\imw]{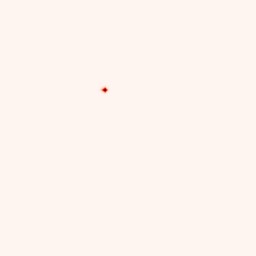} \\

& Input Image & GT mask & QFCA+ & FCA & GRNR & ZvM \cite{Aota_2023_WACV} & April-GAN~\cite{chen2023zero}
\end{tabular}
\caption{\label{fig:qualitative}Visualization of anomaly localization results on challenging examples by different methods. 
}
\end{figure*}

\begin{table}
\centering
\begin{tabular}{l|c|c|c|c}
 & PRO & AUROC$_s$ & $\text{F}_1$ & Time(s)\\
\hline

WinCLIP~\cite{jeong2023winclip} & 71.50 & 89.06 & 38.42 & 0.389 \\
SAA+~\cite{cao2023segment} & 64.79 & 77.82 & 59.19 & 0.270 \\
April-GAN~\cite{chen2023zero} & 92.56 & 96.51 & 58.63 & 0.122 \\ 
SDP+~\cite{chen2024clip} & 92.48 & 96.76 & 52.70 & 0.045 \\ 
AdaptClip~\cite{gao2025adaptclip} & \-- & 97.68 & 55.86 & 0.162 \\
CRANE~\cite{salehi2025crane} & 95.18 & 97.96 & 57.02 & \-- \\
\hline
ZvM~\cite{Aota_2023_WACV} & 93.82 & 97.47 & 60.60 & 1.100 \\ %
FCA\cite{ardelean2023highfidelityArxiv} & 97.18 & 98.73 & 71.75 & 1.070 \\ %
GRNR~\cite{yao2024global} & 94.65 & 97.44 & 62.85 & 0.145 \\
\hline
QFCA$_{512}$ & 97.08 & 98.77 & 68.91 & \textbf{0.014} \\
QFCA (Ours) & 97.13 & 98.72 & 71.88 & 0.057 \\
QFCA+ (Ours) & \textbf{97.57} & \textbf{98.83} & \textbf{73.07} & 0.097 \\
\end{tabular}
\caption{\label{tab:mvtec_all}%
  Results on the MVTec~AD dataset. QFCA$_{512}$ denotes our method ran at a smaller resolution of 512\texttimes{}512. For the other methods we use the image size suggested by their authors.
}
\end{table}

We compare our QFCA/QFCA+ with several zero-shot anomaly detection approaches, including methods based on VLMs: WinCLIP~\cite{jeong2023winclip}, SAA~\cite{cao2023segment}, AprilGAN~\cite{chen2023zero}, and SDP~\cite{chen2024clip}, as well as texture-specific methods: ZvM~\cite{Aota_2023_WACV}, FCA~\cite{ardelean2023highfidelityArxiv}, and GRNR~\cite{yao2024global}.
The results are compared quantitatively in Tab.\ref{tab:mvtec_all} and qualitatively in Fig.~\ref{fig:qualitative}.
It can be observed that QFCA performs significantly better than VLM-based methods and is on par with FCA while being more than 10 times faster. 
Thanks to our feature preprocessing technique, QFCA+ consistently yields the best metrics while still maintaining interactive rates.

We evaluate our method on two additional datasets in Tab.~\ref{tab:quant2}. 
The results suggest that QFCA is versatile and the improvements generalize across various texture classes.
The improvement brought about by our feature preprocessing (QFCA+) is more significant here compared to the MVTec dataset, since the textures are more elaborate.
The benefit of QFCA+ can be observed in Fig.~\ref{fig:qualitative} as a reduction of false positives.

\subsection{Image and patch size}
\label{chap:img_patch_size}

\begin{table}
\centering
\begin{tabular}{l|c|c|c|c}
& PRO & AUROC$_s$ & $\text{F}_1$ & Time(s) \\
\hline
\multicolumn{5}{l}{} \\[-1em]
\multicolumn{5}{l}{WFT (512\texttimes{}512) }\\
\multicolumn{5}{l}{} \\[-1em]
\hline
April-GAN~\cite{chen2023zero} & 84.97 & 94.90 & 71.51 & 0.122 \\ 
ZvM \cite{Aota_2023_WACV} & 84.59 & 96.11 & 72.07 & 1.100 \\
GRNR~\cite{yao2024global} & 80.25 & 96.08 & 73.23 & 0.045 \\
FCA~\cite{ardelean2023highfidelityArxiv} & 89.57 & 98.26 & 79.13 & 0.179 \\
FCA + \knn~\cite{ardelean2023highfidelityArxiv} & 88.77 & 97.73 & 76.30 & 6.345 \\
QFCA (ours) & 89.47 & 98.24 & 78.87 & \textbf{0.013} \\
QFCA+ (ours) & \textbf{92.99} & \textbf{98.51} & \textbf{79.71} & 0.041 \\
\hline
\multicolumn{5}{l}{} \\[-1em]
\multicolumn{5}{l}{ DTD-Synthetic (320\texttimes{}320)}\\
\multicolumn{5}{l}{} \\[-1em]
\hline
April-GAN~\cite{chen2023zero} & 88.50 & 95.32 & 52.40 & 0.075 \\ 
AdaptClip~\cite{gao2025adaptclip} & \-- & 97.70 & 63.60 & 0.162 \\
ZvM \cite{Aota_2023_WACV} & 94.32 & 98.00 & 65.96 & 1.100 \\
GRNR~\cite{yao2024global} & 92.07 & 97.29 & 61.14 & 0.042 \\
FCA~\cite{ardelean2023highfidelityArxiv} & 94.82 & 98.14 & 68.75 & 0.008 \\
FCA + \knn~\cite{ardelean2023highfidelityArxiv} & 95.93 & 98.51 & \textbf{71.79} & 3.970 \\
QFCA (Ours) & 94.90 & 98.14 & 68.97  & \textbf{0.005} \\
QFCA+ (Ours) & \textbf{96.64} & \textbf{98.74} & 71.57 & 0.029 \\

\end{tabular}
\caption{\label{tab:quant2}
  Results on the DTD-Synthetic and Woven Fabric Textures (WFT) datasets.
}
\end{table}

In this section, we study the interplay between image size and patch size, which are paramount to texture anomaly detection.
The size of the input has a strong effect on the resulting scores: while higher resolution images allow for a more precise localization, the receptive field of each feature becomes smaller, reducing the discriminatory potential.
Moreover, as each pixel responds to a smaller area, there is a rising cost in execution time.
We note that generally the patch size used by the anomaly detection method should be scaled proportionally to the image size. This relation is intuitive, since a decrease in feature receptive field can be compensated for by a larger patch size when computing local statistics.

\begin{figure}
	\centering
	\includegraphics[width=1.0\columnwidth]{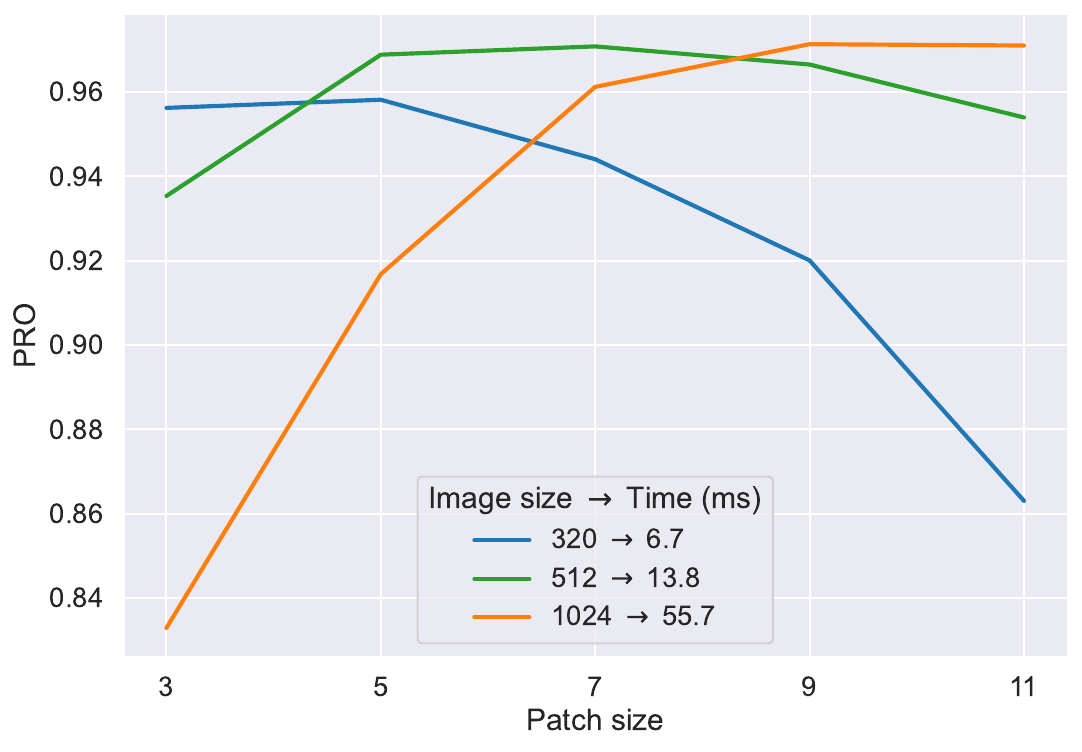}
	\caption{\label{fig:image_patch_size_pro_perf}%
    Diagram showing the \ac{pro} score and execution time with patch sizes 3, 5, 7, 9, or 11. The optimum patch-size depends on the image size.}
	
\end{figure}

Figure \ref{fig:image_patch_size_pro_perf} shows the optimal patch size for different image sizes and their corresponding performance in milliseconds and \ac{pro} score.
The largest image size allows for the highest accuracy, at the cost of increasing computational complexity. 
We empirically observe that the optimal patch size is about $10\%$ of the size of the feature maps, \ie, $1.25\%$ of the size before feature extraction.

\subsection{In-depth ablation study}
In this section we analyze the role and behavior of the newly introduced components and their hyperparameters.

\subsubsection{Number of Bins}
\label{subsec:bin_nr}

\begin{figure}
	\centering
	\includegraphics[width=1.0\columnwidth]{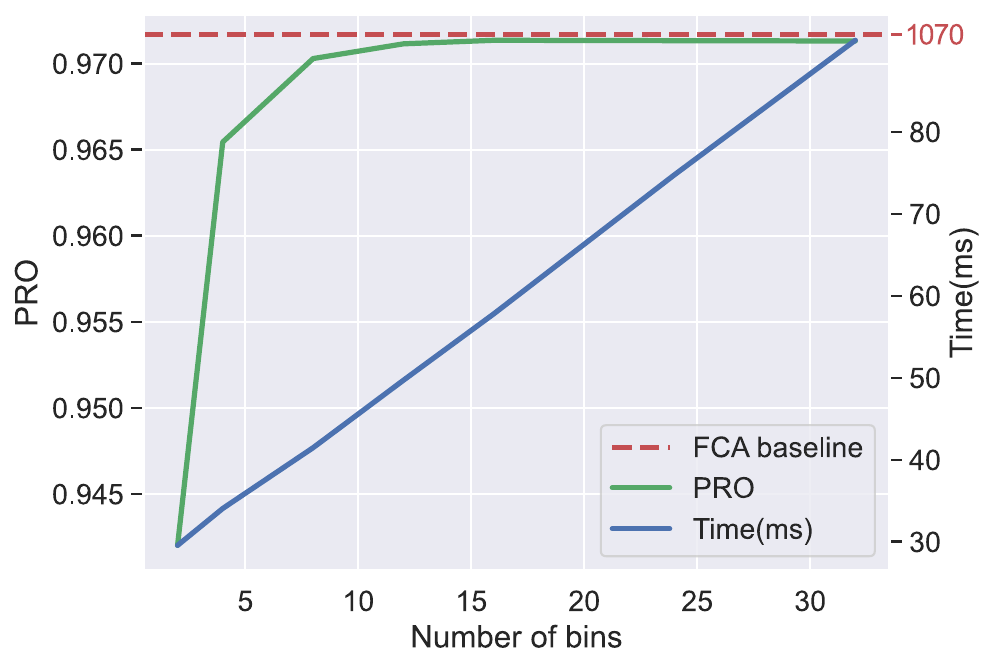}
	\caption{\label{fig:bins_pro_perf}%
    Evaluating the latency and PRO metric of QFCA with increasing number of histogram bins; saturating around 16 bins.}
\end{figure}

A key parameter of \ac{qfca} is the number of bins used in the quantization step.
As described in the method section, increasing the number of bins makes the result closer to the sorting operation in \ac{fca}.
A large number of bins would also hinder the overall performance; however, as seen in Fig.~\ref{fig:bins_pro_perf}, 16 bins are already enough to match the result of \ac{fca}. Note that there is still a very small gap (0.05\% difference) even when using a large number of bins; this is due to differences in postprocessing, as discussed in Sec.~\ref{chap:sigma}.

\subsubsection{Feature preprocessing}
\label{chap:feature_enhancer}

\begin{figure}
	\centering
	\includegraphics[width=0.98\columnwidth]{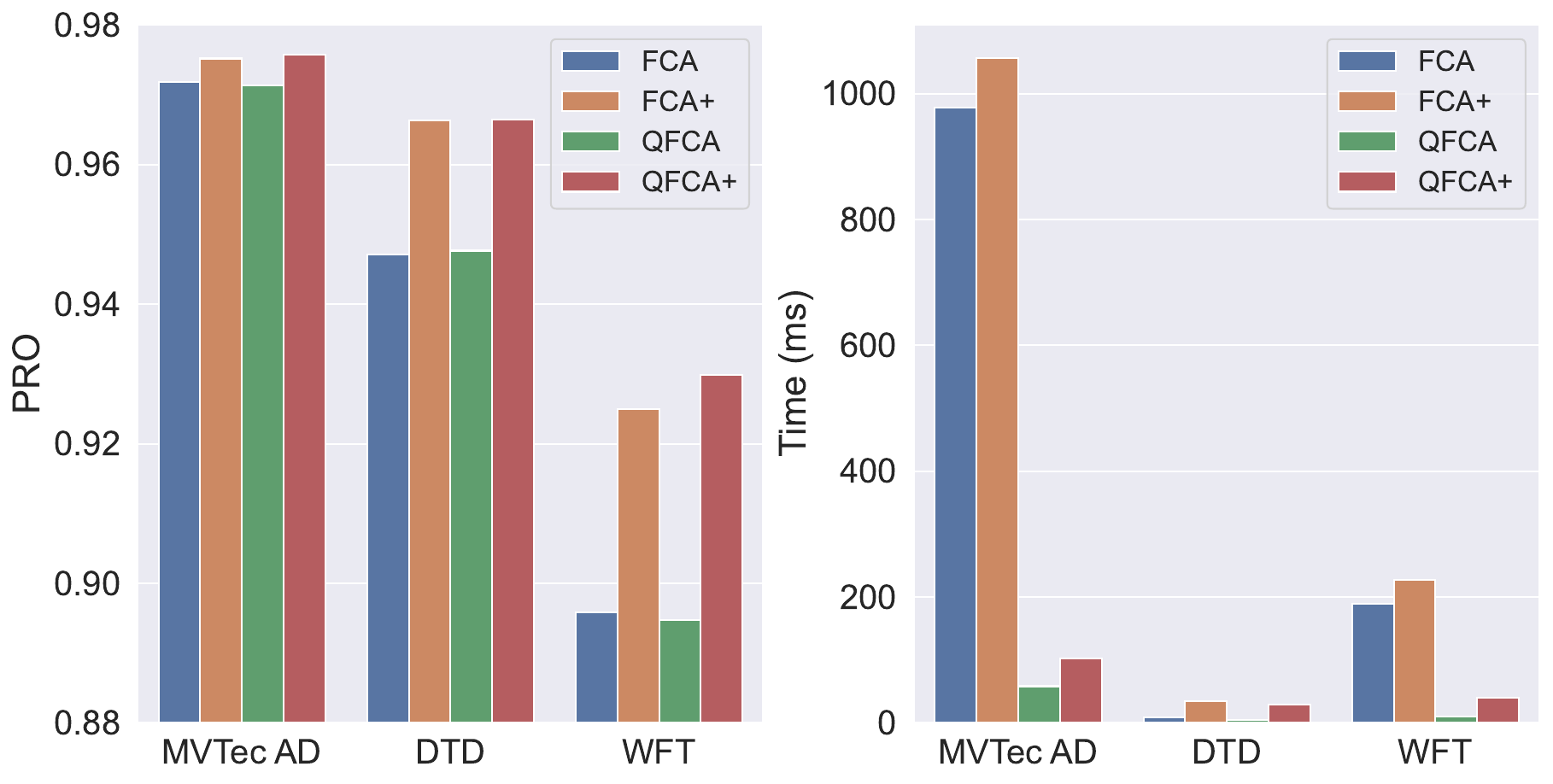}
	\caption{Using our feature preprocessing in conjunction with FCA and QFCA consistently improves the localization performance at the cost of a slightly increased runtime. }
	\label{fig:fe_ablation}
\end{figure}

We ablate our feature preprocessing in Fig.~\ref{fig:fe_ablation}, including the results on different datasets. We show that our feature preprocessing can also be applied directly to FCA, similarly improving its performance (FCA+); in general, the method can be used in conjunction with arbitrary anomaly detection methods.
Depending on the field of application, using our feature preprocessing can provide a substantial improvement, at the cost of a slightly increased running time; that is, about 40 ms for a 1024\texttimes{}1024 image.

\begin{figure}
	\centering
	\includegraphics[width=1.0\columnwidth]{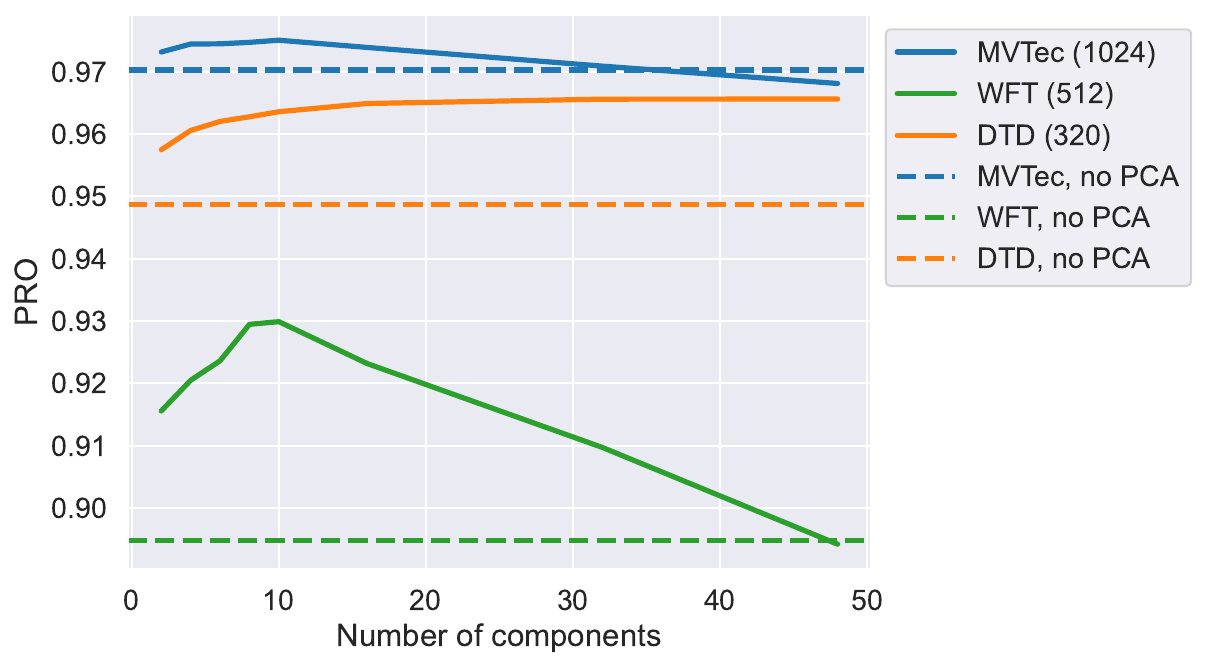}
	\caption{Visualizing the impact of the number of principal components used in preprocessing in terms of \ac{pro} score. Dashed lines show the baseline values, without feature preprocessing.
    }
	\label{fig:fe_enh_n_comp}
\end{figure}

We study the influence of the number of principal components used for feature preprocessing in Fig.~\ref{fig:fe_enh_n_comp}.
Naturally, when using a very small number of components, PCA can scarcely reconstruct the input features, and the residuals do not properly reflect the rarity of the features.
On the other hand, too many principal components allow even infrequent features to be reconstructed well, annulling the benefit of our preprocessing.
We observe that the optimum number of components depends on the complexity of the images (\ie, larger size in MVTec, and intricate textures in DTD-Synthetic).
Nonetheless, 10 principal components proved to be a robust off-the-shelf choice in our experiments.

\subsubsection{Computing the Reference}
\label{chap:reference}
\begin{table}
    \centering
    \begin{tabular}{l|c|c|c}
        \toprule
        Method & PRO &  AUROC$_s$ &  F$_1$ \\
        \midrule
        quan $\rightarrow$ mean   & 96.9 & 98.6 & 71.3 \\
        quan $\rightarrow$ median & 97.0 & \textbf{98.7} & 71.6 \\
        mean $\rightarrow$ quan   & 97.0 & 98.6 & 71.3 \\
        median $\rightarrow$ quan & \textbf{97.1} & \textbf{98.7} & \textbf{71.9} \\
        \bottomrule
    \end{tabular}
    \caption{Comparing different reference selection methods.}
    \label{table:ref_sel}
\end{table}

The reference set $R(F(I))$ describes the global properties across all patches, and it is compared with all patches during error computation (Sec.~\ref{subsec:fca_sc}).
There are multiple ways to determine this reference tensor; the FCA paper tested random selection, computing the mean, computing the median, and \knn. 
The median was found to produce the best results in the case of a single reference and \knn to work best when using multiple references.
Since our work prioritizes execution time, we only consider a single reference $F_r$.

\ac{qfca} has a quantization step before creating the histograms that represent feature patches.
The mean or median of the patch representations can be computed before or after the feature quantization.
Table \ref{table:ref_sel} presents the results obtained with different reference selection procedures.

\subsubsection{Blurring and Averaging Steps}
\label{chap:sigma}

Similarly to FCA, we use two different Gaussian blurring operations as part of the \ac{qfca} algorithm.
The first Gaussian operation, denoted $G_{\sigma_p}(x, y)$ performs the same operations as the spatial weighing function in \ac{fca}.
That is, it aggregates the contribution to error in the patch statistics comparison from the corresponding histogram bin of all surrounding patches. Therefore, the kernel size of the Gaussian filter must be equal to the patch size.

\begin{figure}
	\centering
	\includegraphics[width=0.99\columnwidth]{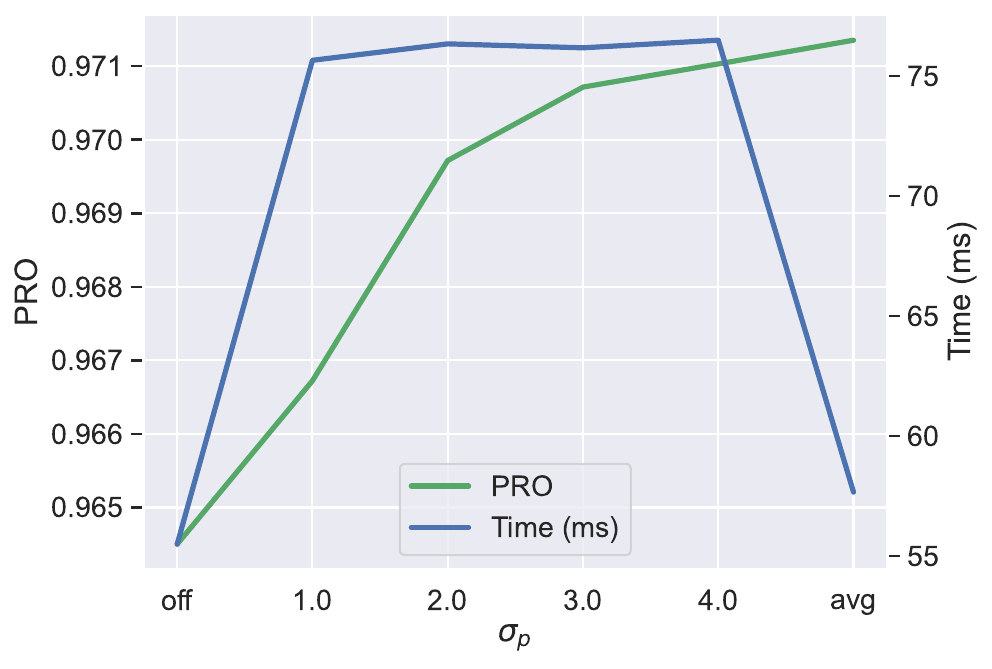}
	\caption{Analyzing the effect of $\sigma_p$ for spatially weighing the mismatch scores. Experiment on the MVTec dataset at 1024\texttimes{}1024.}
	\label{fig:sigma_p_pro_perf}
\end{figure}

Figure \ref{fig:sigma_p_pro_perf} shows a range of possible values for $\sigma_p$. 
The plot reports the results on the MVTec dataset for clarity; however, the DTD-Synthetic dataset and WFT follow the same trend.
It can be seen that the accuracy increases with $\sigma_p$ and peaks when using a simple average pooling, corresponding to $\sigma_p=\infty$, given the fixed kernel size.
This suggests that all contexts (patches) a pixel is part of are equally important.

\begin{figure}
	\centering
	\includegraphics[width=0.99\columnwidth]{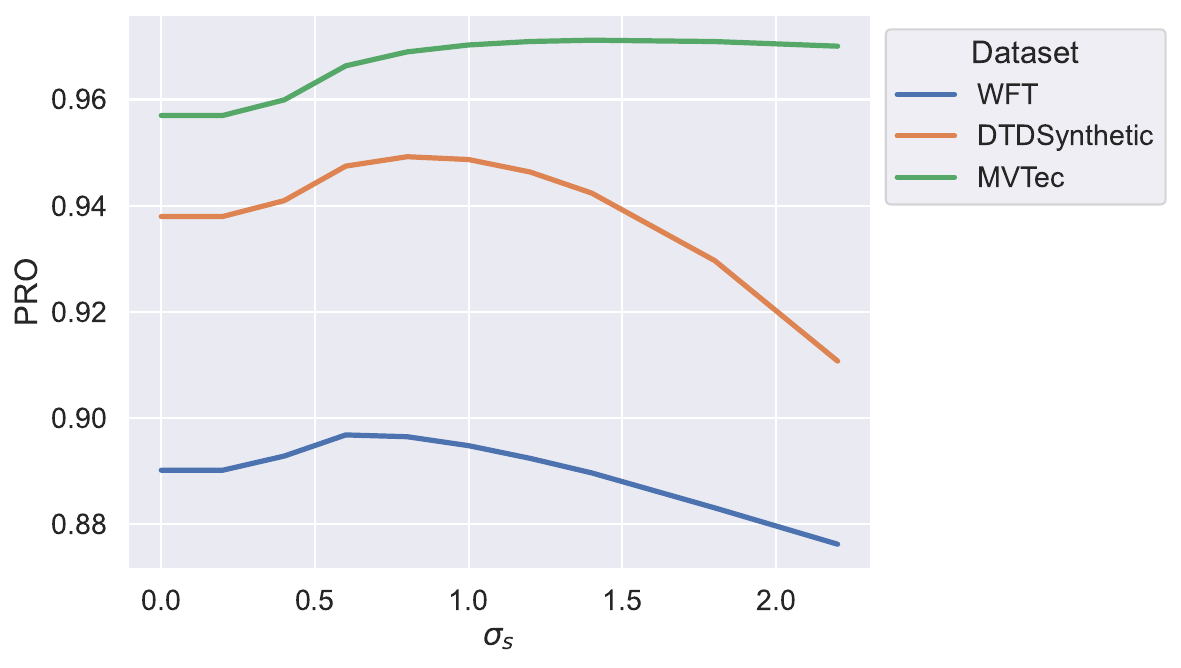}
	\caption{The effect of the final low-pass filter on the PRO score. The running time of this operation is negligible.}
	\label{fig:sigma_s_pro_perf}
\end{figure}

The last step of our algorithm is a 2D blurring step, with standard deviation $\sigma_s$, over the aggregated error contribution scores.
We use the same name $\sigma_s$ as in \ac{fca} because this parameter has the same role as the Gaussian filtering $\mathcal{G}_{\sigma_s}(\cdot)$ used there.
Although not mathematically equivalent, our final blurring serves the same purpose and is significantly more efficient to compute. 
We show the impact of our $\sigma_s$ in Fig.~\ref{fig:sigma_s_pro_perf}.
Based on these observations, we choose $\sigma_s=1$ as our default parameter.

\section{Limitations}
As an optimized algorithm for zero-shot texture anomaly localization, our approach inherits the limitations of this class of methods.
Namely, as we do not use a large VLM to inject general knowledge about anomalies, QFCA is suitable for texture-like data, and not for arbitrary objects.
Nonetheless, thanks to the feature preprocessing, our method brings a significant boost even in this case, compared to \ac{fca}.
Please see the supplementary (\ref{asec:objects}) for an evaluation on such non-textural images.

\section{Conclusion}

In this work, we introduce QFCA+, a real-time anomaly localization method that is not only faster than existing zero-shot approaches, but also detects anomalies with higher precision. More precisely, as seen in Fig.~\ref{fig:teaser}, our QFCA+ offers a more advantageous tradeoff between localization fidelity and runtime.
By replacing the slow sorting operation in \ac{fca} with a GPU-optimized implementation on histograms, we obtain a 10\texttimes{} speedup. Notably, the quantization is able to match the full-precision metrics using just 16 bins.
Moreover, we further speed up the method using a fast average pooling algorithm, with a time complexity invariant to kernel size, which can potentially be included in other pipelines to gain similar speedup benefits.

We additionally propose a feature preprocessing step (the + in QFCA+), which can be used to increase the accuracy of anomaly localization when a slower latency can be afforded.
This addition is beneficial for complex textures and can be used in conjunction with other anomaly detection algorithms.

\myparagraph{Acknowledgements.}
This project has received funding from the European Union’s Horizon 2020 research and innovation programme under the Marie Skłodowska-Curie grant agreement No 956585.

\newpage

\bibliographystyle{eg-alpha-doi} 
\bibliography{main}

\clearpage
\setcounter{page}{1}

\setcounter{section}{0}
\renewcommand\thesection{S\arabic{section}}

\renewcommand{\pageref}[1]{%
  \ifstrequal{#1}{ThisPartLastPage}{3}{\ref{#1}}%
}

\twocolumn[
  \medskip
  \centering{\huge \textbf{Quantized FCA: Efficient Zero-Shot Texture Anomaly Detection}}
  \bigskip
  
  \centering{\large\uppercase{Supplementary Material}}
  \bigskip\medskip
  
  \centering{\large{Andrei-Timotei Ardelean and Patrick Rückbeil and Tim Weyrich}}
  \smallskip
  
  \centering{\normalsize{Friedrich-Alexander-Universität Erlangen-Nürnberg, Germany}}
  
  \bigskip\bigskip%
]

\section{Summary}
In this supplementary material, we include additional visualizations, comparisons, evaluations, and details useful for the reproducibility of our results. 

\section{Detailed quantitative results}
\label{asec:detailed_metrics}

\begin{table*}
\centering
	\begin{tabular}{l|cccc|cccc}
		& \multicolumn{4}{c}{\textbf{QFCA}} & \multicolumn{4}{c}{\textbf{QFCA+}} \\
        \hline
		\textbf{MVTec AD}           & PRO & AUROC$_s$ & F$_1$ & AUROC$_c$ & PRO & AUROC$_s$ & F$_1$  & AUROC$_c$ \\
        \hline
		carpet          & 95.54   & 98.38      & 72.43          & 99.64                                       & 96.72     & 98.77      & 75.75   & 99.48                                              \\
		grid            & 98.16   & 99.50      & 62.29          & 99.84                                      & 98.87     & 99.67      & 67.59   & 100.00                                             \\
		leather         & 98.89   & 99.44      & 65.87          & 99.63                                        & 99.03     & 99.47      & 65.26   & 99.52                                             \\
		tile            & 95.83   & 98.00      & 82.20          & 99.50                                        & 95.88     & 98.02      & 81.12   & 98.19                                              \\
		wood            & 97.24   & 98.27      & 76.59          & 99.30                                        & 97.37     & 98.22      & 75.65   & 97.63                                            \\
        \hline
		\textbf{DTD}             & PRO & AUROC$_s$ & F$_1$ & AUROC$_c$ & PRO & AUROC$_s$ & F$_1$ & AUROC$_c$ \\
        \hline
		Blotchy\_099    & 97.21   & 99.42      & 76.21                              & 99.63                   & 97.22   & 99.13      & 68.82 & 99.94                                                 \\
		Marbled\_078    & 97.17   & 99.05      & 73.45                                 & 100.00                  & 97.32   & 99.00      & 71.49 & 100.00                                               \\
		Mesh\_114       & 94.08   & 97.63      & 63.85                              & 95.80                   & 96.83   & 98.60      & 66.23 & 97.74                                               \\
		Stratified\_154 & 98.78   & 99.13      & 64.87                              & 100.00                  & 98.83   & 99.10      & 63.10 & 100.00                                             \\
		Woven\_068      & 96.98   & 98.79      & 68.70                              & 99.88                   & 97.57   & 99.02      & 70.85 & 100.00                                            \\
		Woven\_125      & 98.08   & 99.37      & 74.85                               & 100.00                  & 98.23   & 99.38      & 74.79 & 100.00                                           \\
		Fibrous\_183    & 97.06   & 99.01      & 71.02                                & 97.75                   & 97.40   & 99.04      & 68.99 & 99.56                                             \\
		Matted\_069     & 89.80   & 99.41      & 74.65                              & 100.00                  & 90.53   & 99.56      & 74.67 & 100.00                                            \\
		Perforated\_037 & 94.88   & 96.82      & 66.00                            & 98.88                   & 96.75   & 97.99      & 67.93 & 99.94                                                 \\
		Woven\_001      & 96.70   & 99.30      & 65.04                               & 96.00                   & 98.57   & 99.69      & 67.53 & 98.38                                            \\
		Woven\_104      & 90.85   & 97.25      & 66.65                             & 95.00                   & 93.92   & 98.10      & 68.46 & 98.94                                         \\
		Woven\_127      & 87.23   & 92.55      & 62.31                                 & 97.11                   & 94.90   & 96.05      & 70.60 & 98.11                                                  \\
        \hline
		\textbf{WFT}             & PRO & AUROC$_s$ & F$_1$ & AUROC$_c$ & PRO & AUROC$_s$ & F$_1$ & AUROC$_c$ \\
        \hline
		texture\_1      & 92.12   & 97.91      & 79.81                                & --                  & 94.25   & 98.06      & 79.68 & --                                             \\
		texture\_2      & 86.83   & 98.57      & 77.92                             & --                  & 91.73   & 98.95      & 79.73 & --                             \\    
               
	\end{tabular}
    \caption{\label{tab:detailed_scores}Per-texture detailed results of our method (QFCA and QFCA+).}
\end{table*}

In Table~\ref{tab:detailed_scores} we include a detailed breakdown of our scores for each texture class in the evaluated datasets.

As mentioned in the main text, the image-level AUROC$_c$ is saturated on the MVTec AD textures. Since the image-level anomaly score is computed as the maximum across the image, the result is quite sensitive to the final smoothing of the results. Table~\ref{tab:aurocc} reports the AUROC$_c$ metric for QFCA and QFCA+ at different smoothing levels $\sigma_s$. In the best case (QFCA+ with $\sigma_s=2.0$) the metric is very close to 100\%. To be exact, the anomalies are perfectly detected in 3 out of 5 textures (grid, leather, and wood).

\section{Evaluation on generic objects}
\label{asec:objects}

\begin{table*}
\centering
    \begin{tabular}{l|cccc|cccc}
    & \multicolumn{4}{c}{\textbf{QFCA+}} & \multicolumn{4}{c}{\textbf{FCA}} \\
    \hline
    \textbf{MVTec AD} & PRO & AUROC$_s$ & F$_1$  & AUROC$_c$ & PRO & AUROC$_s$ & F$_1$ & AUROC$_c$ \\
    \hline
    bottle & \textbf{44.58} & \textbf{56.53} & \textbf{21.27} & \textbf{47.62} & 26.53 & 42.46 & 17.36 & 28.73 \\
    capsule & \textbf{81.57} & \textbf{87.64} & \textbf{13.39} & \textbf{49.30} & 70.38 & 85.34 & 10.48 & 35.14 \\
    grid & \textbf{98.87} & \textbf{99.67} & \textbf{67.59} & \textbf{100.00} & 98.07 & 99.46 & 61.62 & 99.84 \\
    leather & \textbf{99.03} & \textbf{99.47} & 65.26 & 99.52 & 98.90 & \textbf{99.45} & \textbf{66.06} & \textbf{99.63} \\
    metal\_nut & \textbf{39.47} & 55.63 & 35.80 & \textbf{67.89} & 26.38 & \textbf{61.00} & \textbf{36.41} & 59.58 \\
    tile & \textbf{95.88} & \textbf{98.02} & 81.12 & 98.19 & 95.41 & 97.84 & \textbf{81.70} & \textbf{99.36} \\
    transistor & \textbf{40.95} & 59.72 & 16.65 & \textbf{34.92} & 39.13 & \textbf{62.35} & \textbf{17.63} & 30.42 \\
    zipper & \textbf{50.55} & \textbf{85.61} & \textbf{17.09} & \textbf{81.20} & 42.17 & 82.60 & 16.21 & 43.96 \\
    cable & 26.64 & 70.02 & 14.23 & \textbf{49.16} & \textbf{32.84} & \textbf{75.22} & \textbf{26.88} & 45.46 \\
    carpet & \textbf{96.72} & \textbf{98.77} & \textbf{75.75} & 99.48 & 95.44 & 98.31 & 72.58 & \textbf{99.60} \\
    hazelnut & \textbf{93.72} & \textbf{92.67} & \textbf{56.71} & \textbf{97.82} & 91.08 & 91.78 & 50.31 & 95.00 \\
    pill & \textbf{78.14} & 77.02 & 22.17 & \textbf{52.48} & 76.08 & \textbf{80.84} & \textbf{25.62} & 45.39 \\
    screw & \textbf{86.37} & \textbf{96.08} & \textbf{16.39} & 50.46 & 77.85 & 93.39 & 5.94 & \textbf{53.93} \\
    toothbrush & \textbf{82.88} & \textbf{92.09} & \textbf{31.29} & \textbf{91.11} & 64.26 & 87.20 & 16.22 & 73.06 \\
    wood & \textbf{97.37} & \textbf{98.22} & 75.65 & 97.63 & 97.18 & \textbf{98.22} & \textbf{76.34} & \textbf{99.30} \\
    \hline
    Average & \textbf{74.18} & \textbf{84.48} & \textbf{40.69} & \textbf{74.45} & 68.78 & 83.70 & 38.76 & 67.23 \\
    
    \end{tabular}
    \caption{\label{tab:obj_scores}Per-class results of our QFCA+ compared to FCA on all objects in the MVTec AD dataset.}
    \end{table*}

\begin{table}[!t]
    \centering
    \begin{tabular}{l|ccccc}
         & \multicolumn{5}{c}{$\sigma_s$}\\
         \hline
        Method & 1.2 & 1.4 & 1.6 & 1.8 & 2.0 \\
        QFCA & 99.83 & 99.83 & 99.82 & 99.82 & 99.80 \\
        QFCA+ & 99.38 & 99.70 & 99.77 & 99.84 & \textbf{99.89}
    \end{tabular}
    \caption{Average image-level AUROC$_c$ on MVTec AD textures at different levels of smoothing.}
    \label{tab:aurocc}
\end{table}

\begin{table}
    \begin{tabular}{lccc}
        Method & PRO & AUROC$_s$ & F$_1$ \\
        \hline
        WinCLIP~\cite{jeong2023winclip} & 64.6 & 85.1 & 31.7 \\
        April-GAN~\cite{chen2023zero} & 44.0 & 87.6 & 43.3 \\
        AnoVL~\cite{deng2023anovl} & 77.8 & 90.6 & 36.5 \\
        SDP~\cite{chen2024clip} & 79.1 & 88.7 & 35.3 \\
        SDP+~\cite{chen2024clip} & 85.6 & 91.2 & 41.9 \\
        SAA~\cite{cao2023segment} & 31.9 & 67.7 & 23.8 \\
        SAA+~\cite{cao2023segment} & 42.8 & 73.2 & 37.8 \\
        ClipSAM~\cite{li2025clipsam} & 88.3 & 92.3 & 47.8 \\
        \hline
        QFCA+ (Ours) & 74.2 & 84.5 & 40.7
    \end{tabular}
\caption{\label{tab:VLM_scores}Comparison to VLM-based methods on all objects in MVTec AD; metrics taken from~\cite{li2025clipsam}. While our method is specific to textures, it is on par with most VLM-based methods that make use of significantly more pretraining as well as textual clues.}
\end{table}

Our method is designed to work on textures or surfaces that are largely stationary. 
While that is the case, our feature preprocessing in QFCA+ significantly improves the performance on images that are just partly texture-like.
Table \ref{tab:obj_scores} includes an evaluation on all objects in MVTec in a comparison with FCA.
For some classes (such as transistor) the difference between QFCA+ and FCA is rather small. However, objects that can be characterized as two intertwined textures (such as screw and toothbrush) see large improvements since they benefit most from our PCA-based preprocessing.
Moreover, as seen in Tab.~\ref{tab:VLM_scores}, despite being designed for textures and using limited pretraining (WideResnet on ImageNet), QFCA+ outperforms WinCLIP~\cite{jeong2023winclip} and is comparable to more recent VLM-based methods. 

\section{Proof of algorithm correctness}
\label{ssec:correct}

In this section, we prove the correctness of the algorithm introduced for computing the mismatch score between a patch and a reference histogram (Alg. \ref{algorithm}).
In this context, correctness means that the algorithm is equivalent to the feature correspondence mismatch from FCA, \ie $M(x, y; P)$ in ~\cite{ardelean2023highfidelityArxiv}. Specifically, the computed errors for each bin match the errors obtained using the FCA algorithm if it were run on the quantized values.

For simplicity, we analyze the algorithm for a single patch, assume integer weights, and use the same notation as in Alg. \ref{algorithm}. Additionally, let $\{X_i\}_{i=1}^{T^2}, \{Y_i\}_{i=1}^{T^2}$ be the feature values of the patch and the reference, respectively.
FCA computes the mismatch score by sorting the $X$ and $Y$ vectors and mapping elements of the same rank. The error associated with each element of a patch is given by the absolute difference to its matched value; \ie, $\texttt{FCA}_k := |X_{O_k} - Y_{L_k}|$, where $O$ and $L$ are the indices of the sorted order of $X$ and $Y$.
For the case where the histogram weight of a bin is one, \ie $P_i = 1$, it is easy to see that $E_i = |Q_i - Q_j|$, as per lines 6 or 10 of Alg.~\ref{algorithm}. Therefore, the error $E_i$ corresponds to $\texttt{FCA}_k$ if and only if $i$ and $j$ correspond to the same ordered rank.
$P$ and $R$ represent the vectors as histograms, where the bins correspond to quantiles $Q$, which are inherently in sorted order. It follows that $i$ and $j$ track the ranks in sorted order iff at each iteration the cumulative weight of the processed bins in histogram $P$ matches the cumulative weight in histogram $R$; that is, for the order statistic $k$, $\sum_{b=1}^{j-1}R_b < k \le \sum_{b=1}^{j}R_b$ and $\sum_{b=1}^{i-1}P_b < k \le \sum_{b=1}^{i}P_b$. 
Since we process the order statistics for one bin at a time, this translates in our algorithm to
$\sum_{b=1}^{j-1}R_b < \sum_{b=1}^i P_b \le \sum_{b=1}^j R_b$ if the algorithm enters the condition on line 5 and $\sum_{b=1}^{i-1}P_b < \sum_{b=1}^j R_b \le \sum_{b=1}^i P_b$ otherwise.

It can be verified by induction that these identities hold. At each iteration, the cumulative sum for both histograms is adjusted by the same amount: the index of the smaller bin is moved forward and the discarded weight is taken from the larger bin to mark the transport.
Since the weights are nonnegative, the new cumulative sum is bounded by the interval defined by the bins of the other histogram.
This remark is sufficient for the case where $P_i = 1$; however, in practice the elements will not be unique due to quantization.
When the vectors $X$ and $Y$ have duplicate values, the FCA mismatch score for these elements is ill-defined, since different sorted orders are possible. 
Our QFCA algorithm alleviates this issue by defining the mismatch score of a bin as the weighted average of the scores pertaining to a bin. 
The multiplication with $P_i$ on line 6 (and $R_j$ on line 10, respectively) is balanced by the division by $\hat{P}_i$ on line 15 to create this weighted average.
Intuitively, in the context of FCA, this translates to computing the mismatch scores in sorted order as before, and then averaging the errors of elements with the same value (because they are arbitrarily mapped to different values in the reference).

We have thus shown that the histogram-based algorithm yields the same mismatch scores as the standard FCA used on quantized values. Note that this also implies that the quantized algorithm exactly converges to the patch statistics comparison of FCA as the number of bins goes to infinity.

\myparagraph{2-Wasserstein distance}
We provide here an additional mathematical justification for the efficacy of FCA and QFCA for localizing anomalies.
Ardelean and Weyrich~\cite{ardelean2023highfidelityArxiv} present the patch comparison algorithm as an ad-hoc trick to obtain the contribution of each element to the 1-Wasserstein distance.
That being said, we show here that using the absolute value between matching elements in sorted order is not arbitrary, but rather it represents the magnitude of the gradient of the squared 2-Wasserstein distance between the two distributions.

The Wasserstein distance for distributions with dimensionality $d=1$ has the analytic solution:
\begin{align}
    W_p(X, Y) = \left(\int_0^1 |F^{-1}(z) - G^{-1}(z)|^p dz\right)^{1/p}  \;,
\end{align}
where $F$ and $G$ are the cumulative distributions of $X$ and $Y$, respectively.
For our empirical distributions this can be written as
\begin{align}
    W_p(X, Y) = \left(\sum_{k=0}^{T^2} |X_{O_k} - Y_{L_k}|^p \right)^{1/p}  \;.
\end{align}

The squared 2-Wasserstein distance is then simply
\begin{align}
    W^2_2(X, Y) = \sum_{k=0}^{T^2} (X_{O_k} - Y_{L_k})^2  \;,
\end{align}
so the magnitude of the gradient of an element $X_{O_k}$ is given by
\begin{align}
    \left|\frac{\partial W_2^2}{\partial X_{O_k}}\right| = 2|X_{O_k} - Y_{L_k}| = 2\cdot\texttt{FCA}_k  \;.
\end{align}
Since multiplying all scores by a constant factor does not change the anomaly localization, it follows that QFCA can be seen as efficiently computing the gradient of the squared 2-Wasserstein distance between all patches and the global reference.

\section{Additional Details}
All time measurements of our method as well as the baselines of FCA~\cite{ardelean2023highfidelityArxiv}, GRNR~\cite{yao2024global}, ZvM~\cite{Aota_2023_WACV}, and April-GAN~\cite{chen2023zero} were performed using an NVIDIA RTX A5000 GPU. For the other baselines we use the runtime reported by the authors.

For our local average pooling experiment in Figure~\ref{fig:avg_pooling} we use up-to-date versions of the libraries. That is: Tensorflow version 2.18.0, Pytorch version 2.6.0, and Jax version 0.5.2\,.

\section{Code}
The code is available at: \href{https://github.com/TArdelean/QuantizedFCA}{github.com/TArdelean/QuantizedFCA}.

\label{part2end}
\def\partIIlastpage{\pageref{part2end}}

\end{document}